\documentclass[letterpaper, 10 pt, conference]{deps/ieeeconf}

\IEEEoverridecommandlockouts
\usepackage[utf8]{inputenc}
\usepackage[T1]{fontenc}
\usepackage{csquotes}
\usepackage{float}
\usepackage{amsmath,amssymb}
\usepackage{xcolor}
\usepackage{algpseudocodex}
\usepackage{array,tabularx,makecell,booktabs}
\usepackage{pifont}
\usepackage{rotating}
\usepackage{graphicx}

\usepackage[bookmarks=false]{hyperref}
\hypersetup{
    colorlinks=true,
    linktoc=all,
    citecolor=black,
    filecolor=black,
    linkcolor=black,
    urlcolor=black
}

\usepackage{caption}
\usepackage{xspace}

\makeatletter
\DeclareRobustCommand\TODO{\textbf{TODO}\@latex@warning{TODO used}\xspace}
\makeatother

\newcommand{\acaption}[2]{\caption[#1]{\emph{#1}:~#2}}

\newcommand{\cmark}{\textcolor{green!60!black}{\ding{51}}}
\newcommand{\xmark}{\textcolor{red!70!black}{\ding{55}}}
\newcommand{\pmark}{\textcolor{orange!80!black}{$\sim$}}
\usepackage{soul}

\newcommand{\am}[1]{\sethlcolor{cyan}\hl{\textbf{Archan:} #1}}
\renewcommand{\am}[1]{}

\def\WORKNAME{D3D-GEN\xspace}

\title{\LARGE \bf
\WORKNAME: Robot-Aware Domain-Grounded Interactive 3D World Generation for Social Robotics
}

\author{
Do Duc Anh$^{1}$,
Volodymyr Shcherbyna$^{1,2}$,
Duc Tai Nguyen$^{1}$,
Spaarsh Thakkar$^{1}$,\\
Zhengcheng Shen$^{3}$,
Teham Buiyan$^{4}$,
Archan Misra$^{1}$,
Linh K{\"a}stner$^{1}$ \\[0.3em]
{\small $^{1}$Singapore Management University,}
{\small $^{2}$Technical University Berlin,}
{\small $^{3}$Max Planck Institute,}
{\small $^{4}$Technical University Braunschweig}
\\[0.2em]
}

\usepackage{capt-of}

\makeatletter
\let\@oldmaketitle\@maketitle
\renewcommand{\@maketitle}{\@oldmaketitle
  \begin{center}
    \includegraphics[width=0.99\textwidth]{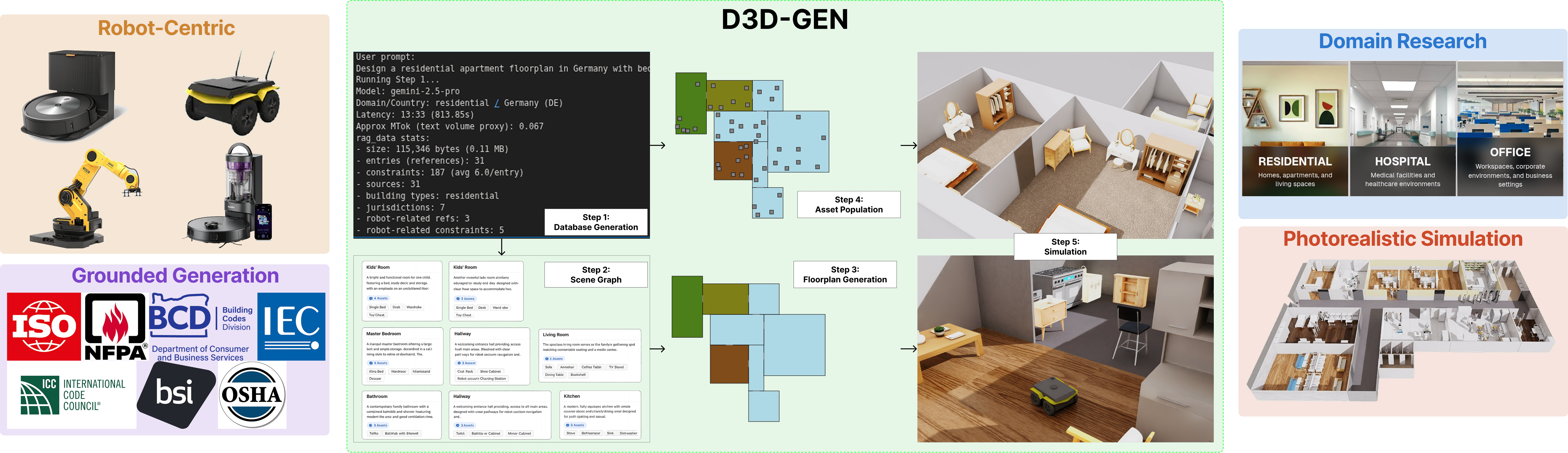}
    \captionof{figure}{\WORKNAME is a novel domain-generalized approach to 3D world generation for embodied AI deployment.
    The system first gathers domain knowledge into a database with legal regulations and conventions, then a user prompt is grounded in the database. A room-level scene graph is constructed with semantic nodes and adjacency edges, along with the preliminary assets of the room. The system then generates room polygons satisfying spatial and connectivity constraints, then places assets according to the rules from the database. This pipeline results in an unlimited amount of realistic, logically sound worlds.
    The resulting world is simulable and can be loaded in Isaac Sim and Gazebo for robotic simulation.}
    
    \label{fig:intro}
  \end{center}
}
\makeatother

\usepackage{makecell}
\usepackage{stfloats}

\usepackage{listings}
\begin{document}

\maketitle

\thispagestyle{empty}
\pagestyle{empty}

\begin{abstract}
Training and validation of Embodied AI for social navigation critically depends on realistic simulation environments, yet many current approaches fail to find a balance between realism and simulability.
We propose \emph{\WORKNAME}, a novel world generation system that combines a domain agent with a retrieval-augmented generation (RAG) pipeline grounded in that domain. Our system enables users to rapidly generate domain-grounded, fully interactive 3D worlds by automating both the collection of domain knowledge and the synthesis of realistic floorplans and object placements, without dependence on any fixed 3D model database.
Given a domain description prompt, the research agent collects publicly accessible domain-specific data and constructs a persistent domain database.
Using this database, our RAG pipeline generates plausible floorplans and object placements by dynamically querying a user-provided semantic database, which can be easily extended or modified.
The output is a fully interactive 3D world loadable by the popular simulators Isaac Sim and Gazebo.
With our approach, we have built databases for several common domains (indoor residential, hospital, office) and generated dozens of distinct, plausible simulation environments for each domain.
We present \WORKNAME with a local web frontend that facilitates rapid, interactive world generation for robot simulation.
\end{abstract}

\section{Introduction}
Robotics research increasingly relies on simulated environments for training and validation.
A constant challenge for simulated approaches is successfully bridging the simulation-to-real (sim2real) gap~\cite{sim2realgap}.
The quality of the simulated 3D environment plays a pivotal role in the real-world transferability of simulation-native models.
Environment diversity, semantic soundness, and perceptual realism directly impact training outcomes.

Agents trained on homogeneous environments develop navigation strategies that generalize poorly to out-of-distribution spaces~\cite{embodied-survey}. Part of the field has converged on a narrow set of training scenarios, predominantly sanitized residential scenes~\cite{3dfront}.
Hospital wards, residential buildings, and commercial real estate follow distinct legal regulations, spatial conventions, and object distributions.
These nuances cannot be captured by a single one-size-fits-all approach.

Closing this gap requires simulation environments that are not only physically interactive, but semantically grounded in the conventions of the specific domain in which the robot will operate.
Current methods that prioritize perceptual realism tend to produce physically unsound scenes, methods that prioritize simulability generate domain-generic layouts that lack semantic soundness. This is critical for robotics. Robotics research needs environments that are operationally viable, not just visually appealing scenes that do not transfer to real deployments. Perceptual realism alone does not ensure adequate working space for agents, especially when working with specific domains like office or hospital. Domain grounding is therefore essential for robot operation.

\begin{table*}[!htp]
\centering
\begin{tabularx}{\textwidth}{l *{9}{>{\centering\arraybackslash}X}}
\toprule
& \multicolumn{2}{c}{\textbf{Realism \& Correctness}}
& \multicolumn{2}{c}{\textbf{Generation}}
& \multicolumn{3}{c}{\textbf{Accessibility}}
& \multicolumn{2}{c}{\textbf{Capabilities}} \\
\cmidrule(lr){2-3}\cmidrule(lr){4-5}\cmidrule(lr){6-8}\cmidrule(lr){9-10}
Method
  & Physical Soundness
  & Semantic Grounding
  & Domain Generality
  & Scalable Generation
  & Simulation Ready
  & Natural Language
  & Variable Assets
  & Generate Floorplan
  & Place Assets
\\
\midrule
\multicolumn{10}{l}{\textit{Procedural generation}} \\
ProcTHOR~\cite{procthor}
 & \cmark & \xmark & \xmark & \cmark & \cmark & \xmark & \cmark & \cmark & \cmark \\
Infinigen~\cite{infinigen-indoor}
 & \cmark & \xmark & \pmark & \pmark & \cmark & \xmark & \xmark & \cmark & \cmark \\
\midrule
\multicolumn{10}{l}{\textit{Diffusion-based generation}} \\
DiffuScene~\cite{diffuscene}
 & \pmark & \xmark & \xmark & \xmark & \cmark & \xmark & \xmark & \xmark & \cmark \\
RoomDreamer~\cite{roomdreamer}
 & \xmark & \xmark & \xmark & \xmark & \xmark & \xmark & \xmark & \xmark & \cmark \\
HouseDiffusion~\cite{housediffusion}
 & \xmark & \xmark & \xmark & \xmark & \xmark & \xmark & \xmark & \cmark & \xmark \\
\midrule
\multicolumn{10}{l}{\textit{Agentic generation}} \\
LayoutGPT~\cite{layoutgpt}
 & \pmark & \xmark & \xmark & \cmark & \cmark & \cmark & \xmark & \xmark & \cmark \\
PhyScene~\cite{physcene}
 & \cmark & \xmark & \xmark & \xmark & \cmark & \xmark & \xmark & \xmark & \cmark \\
SAGE~\cite{sage}
 & \cmark & \xmark & \xmark & \pmark & \cmark & \xmark & \xmark & \xmark & \cmark \\
Holodeck~\cite{holodeck}
 & \pmark & \xmark & \xmark & \cmark & \xmark & \cmark & \cmark & \cmark & \cmark \\
\midrule
\textbf{\WORKNAME (Ours)}
 & \cmark & \cmark & \cmark & \cmark & \cmark & \cmark & \cmark & \cmark & \cmark \\
\bottomrule
\end{tabularx}
\caption{
\centering
World generation approaches that are relevant to indoor scene generation for Embodied AI.\\\cmark~supports, \xmark~does not support, \pmark~partially supports.}
\label{tab:comparison}
\end{table*}

We propose \WORKNAME to address these limitations.
Our system autonomously retrieves and organizes domain-specific knowledge based on a research prompt, grounding layout and object placement decisions in verifiable real-world conventions rather than generic priors.
A subsequent world generation prompt allows both domain experts and laypeople to describe the world the way they imagine it.

The result is a physically plausible, semantically sound 3D environment ready for simulation, generated directly from user intent and grounded in verifiable real-world rules. We also provide generated worlds with semantic annotations, allowing robot agents to be fully integrated into the robot simulation pipeline. 
Our system is designed to scale horizontally with an arbitrary number of parallel workers generating worlds from the same or different prompts, at any pipeline stage.
We also provide a web frontend to provide a user-friendly feedback-driven experience, with editable previews generated after every pipeline stage.

\section{Related Work}

Generating indoor environments for Embodied AI has taken three broad directions, each with its own trade-offs between realism, physical soundness, and scalability.

Real-world scan datasets like HM3D~\cite{hm3d} and Matterport3D~\cite{m3d} capture genuine spatial complexity, but the meshes they produce are non-watertight and require significant manual post-processing before they are compatible with physics engines like Isaac Sim.
These datasets also contain no interactive objects; doors and drawers are static geometry rather than articulated components, limiting their usefulness for social navigation tasks that require object interaction. Platforms like Habitat~\cite{habitat} and AI2-THOR~\cite{ai2thor} wrap such datasets with physics engines, but remain restricted to the domains covered by the datasets.

Designer-created datasets like 3D-FRONT~\cite{3dfront} and Structured3D~\cite{structured3d} offer physically sound CAD assets, but their layouts tend toward the sanitized end of the spectrum, stripped of the clutter and imperfections that characterize real human spaces.
3D-FRONT in particular has become the default dataset for a large body of scene generation work, including DiffuScene~\cite{diffuscene}, InstructScene~\cite{instructscene}, SceneTeller~\cite{sceneteller}, RoomDreamer~\cite{roomdreamer}, and LayoutGPT~\cite{layoutgpt}, yet it has significant limitations.
The dataset is extremely large, with asset libraries running to hundreds of gigabytes, creating a substantial barrier to entry for new domains or deployment contexts.
More importantly, 3D-FRONT covers only indoor residential settings: bedrooms, living rooms, and dining areas. Models trained or evaluated on these data inherit that restriction and do not transfer to hospitals, airports, warehouses, or public spaces where social robots are increasingly expected to operate. Recent simulation platforms like RoboCasa~\cite{robocasa} and HumanTHOR~\cite{humanthor} have begun addressing household manipulation scenarios, but neither supports multi-domain generation from a single pipeline.

Procedural approaches like ProcTHOR~\cite{procthor} and Infinigen~\cite{infinigen-indoor} address scalability and physical soundness, but tend to produce sterile, homogeneous layouts skewed toward simple rectangular rooms. 

Recent agentic and diffusion-based methods represent the current frontier.
PhyScene~\cite{physcene} uses differentiable guidance during generation to regulate object quantity, prevent articulation collisions, and preserve navigable space.
SAGE~\cite{sage} employs a physics critic that detects and iteratively corrects placement failures before a scene is finalized.
LLplace~\cite{llplace} frames layout synthesis as an LLM reasoning task, letting a language model directly propose room arrangements, but it relies on a single-shot generation step without external domain grounding or multi-stage refinement.
Graph2Scene~\cite{graph2scene} generates 3D indoor scenes from interaction-aware scene graphs, yet its graph vocabulary and asset library are fixed to the 3D-FRONT residential domain.
While representing meaningful advances, these methods operate over fixed asset databases and predefined spatial patterns, which limits their ability to produce environments grounded in the specific conventions of a target domain. However, none of these methods incorporate robot-operational constraints as first-class generation objectives, including the infrastructure required for autonomous robot operation. Environments generated by such methods require significant manual effort before they are viable for robot training or benchmarking scenarios.

Parallel work on \emph{scene-graph representations} underscores how structured spatial knowledge can improve robotic reasoning.
Domain-Conditioned Scene Graphs~\cite{domain-conditioned-sg} show that conditioning graph nodes on domain semantics enables more effective state-grounded task planning, while FunGraph~\cite{fungraph} attaches functionality labels to scene-graph nodes so that language-prompted agents can reason about object affordances.
SPADE~\cite{spade} builds actionable multi-domain 3D scene graphs and uses them for scalable path planning.
TACS-Graphs~\cite{tacsgraphs} integrates ground-robot traversability into room segmentation, showing that traversability-aware scene graphs yield more consistent room boundaries and better loop-closure detection—a property directly relevant to the corridor and circulation constraints that \WORKNAME encodes in its generated floorplans.
These methods consume scene graphs but do not generate the underlying 3D worlds; \WORKNAME bridges that gap by producing both the scene graph and the simulator-ready environment from a single natural language prompt.

A convincing hospital ward requires knowing where nurses keep medication trolleys, how wide corridors must be for wheelchair access, and what a busy ward looks like at shift change.
That knowledge is domain-specific, contextual, and not derivable from generic training data.
Existing tools also expose generation through technical parameters, which puts them out of reach of domain experts who lack simulation expertise but have the deepest understanding of target environments.
\WORKNAME addresses both gaps by grounding generation in autonomously retrieved domain knowledge and accepting natural language as its primary interface.

\section{Methodology}
\setcounter{figure}{1}
\begin{figure*}[!ht]
    \includegraphics[width=.99\linewidth]{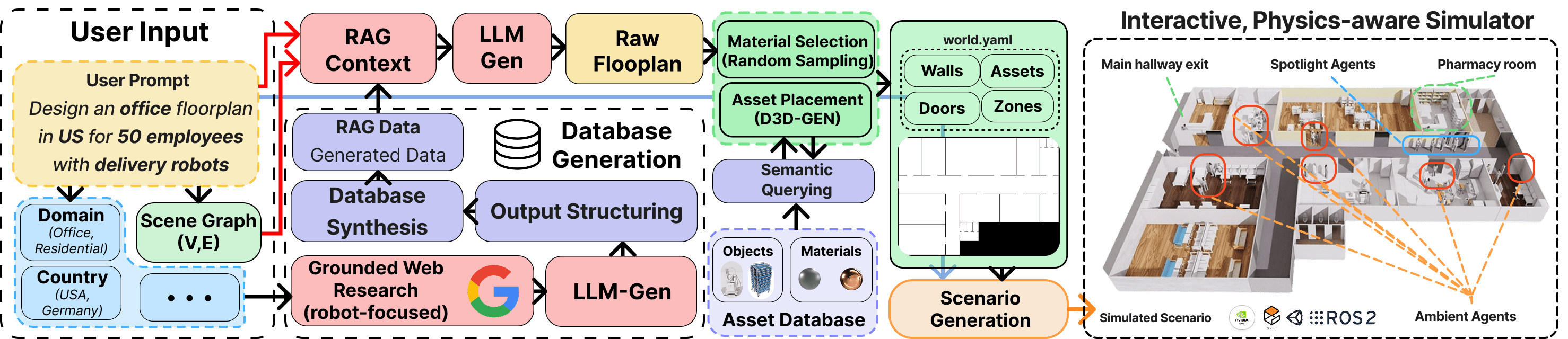} \acaption{System Design}{Overview of the end-to-end \textit{Prompt-to-Simulator Workflow} that transforms a natural language prompt into an executable, physics-aware multi-agent simulation. The pipeline consists of three tightly coupled stages: \emph{database generation} (domain agent), \emph{world generation}, and \emph{scenario generation}. The user provides a prompt together with lightweight structural metadata, including a domain label (Residential, Office, Hospital, \ldots), a jurisdiction/country tag (e.g., USA, Germany), and an optional scene graph encoding room nodes and adjacency edges. A retrieval-augmented generation (RAG) component conditions all downstream generation by constructing a domain-specific knowledge base: grounded, robot-focused web research is synthesized and structured by an LLM into a deduplicated corpus and a finalized \texttt{rag\_data.generated.json} database, enabling semantic querying over constraints, norms, and design priors. Conditioned on this RAG context, the LLM generates a \emph{raw 2D floorplan} that is then compiled into a 3D world through two coupled mechanisms: (i) \emph{material selection} via randomized sampling to diversify physical appearance and surface properties, and (ii) \emph{asset placement} (\WORKNAME) driven by semantic queries over an \emph{asset database} to populate rooms with functionally consistent objects. The resulting world specification (\texttt{world.yaml}) encodes walls, doors, assets, and semantically labeled zones (Regions of Interest, ROIs) that ground both navigation and behavior anchoring. Finally, a \emph{scenario generation} module instantiates agents and task structure over the ROIs, producing an interactive simulator scene with both \emph{spotlight agents} (task-relevant) and \emph{ambient agents} (background), yielding scalable, coherent behaviors aligned with the user intent and the generated environment.}
    \label{img:systemdesign}
\end{figure*}

\subsection{Domain Database}
The \emph{Domain Database} is a structured knowledge base that encodes domain-specific constraints, serving as the bridge between user intent and practical reality.
Rather than providing a static, hand-curated knowledge base, we frame database construction itself as an \emph{autonomous generation task}: given only a single natural-language request, the system synthesizes a schema-validated database of domain-specific constraints by conducting grounded web research, extracting quantitative rules from authoritative sources, and organizing them into a machine-readable format.
Every constraint in the resulting database carries explicit provenance (source URL, publisher, jurisdiction, confidence level), making the knowledge inspectable, citable, and reproducible.

\subsubsection{Database Generation Pipeline}\label{sec:db-gen-pipeline}

The database is produced by a single end-to-end pipeline that accepts a user prompt and proceeds through four sequential stages, each implemented as separate LLM calls:

\begin{enumerate}
  \item \textbf{Domain \& Country Inference.}
    A structured-output call classifies the user prompt into a canonical triple $(\mathit{domain},\,\mathit{country},\,\mathit{country\_code})$, where $\mathit{domain}\in\{\text{residential},\,\text{office},\,\text{hospital}\}$.

  \item \textbf{Grounded Web Research.}
    For each of eight predefined topics, the pipeline issues a Gemini call with Google~Search grounding enabled, repeated for $k$ passes (default $k{=}2$).
    Each call retrieves public standards, codes, guidelines, handbooks, and research documents relevant to the inferred domain and country, returning raw notes with embedded URLs.
    The eight topics are:
    (a)~room types and required support spaces,
    (b)~room size ranges per room type,
    (c)~room arrangements and adjacency rules,
    (d)~fire safety and evacuation,
    (e)~employee/resident convenience and accessibility,
    (f)~energy efficiency and HVAC/day lighting,
    (g)~common objects, furniture, and equipment per room type, and 
    (h)~robot placement, charging, circulation, and docking.
    With $k{=}2$ passes, this stage issues $8\times 2=16$ grounded search calls.

  \item \textbf{Corpus Structuring.}
    The raw notes are split into chunks and each chunk is converted into a list of typed documents following a corpus schema.
    Documents from different chunks that share the same URL are merged: metadata fields are unified and constraints are deduplicated by a composite signature $(\mathit{name},\,\mathit{type},\,\mathit{value})$.

  \item \textbf{Database Synthesis.}
    A deterministic function transforms the corpus into the final database.
    For each unique document it generates a stable reference slug, maps the document type to one of seven canonical source types, normalizes each constraint (validating type, measurement basis, and coercing value into one of four polymorphic forms), and attaches provenance links back to the originating source.
    The output is optionally validated against a schema.
\end{enumerate}

\begin{figure}[t]
\centering
\begin{lstlisting}[basicstyle=\ttfamily\scriptsize,frame=single]
Document
+-- title, publisher, url
+-- doc_type : regulation | guideline | ...
+-- edition_or_year
+-- jurisdiction_context
+-- topics_covered[]
+-- room_types[]
+-- constraints[]
    +-- name, constraint_type, value, unit
    +-- measurement_basis
    +-- applicability : {space_type, conditions[]}
    +-- confidence : high | medium | low
\end{lstlisting}
\caption{Simplified corpus document schema used in Stage~3.}
\label{fig:corpus-schema}
\end{figure}

\subsubsection{Database Schema}\label{sec:db-schema}

The final database JSON conforms to a strict schema with two top-level sections:
(1)~\textbf{Metadata}, recording provenance (generation timestamp, model identity, building-type scope); and
(2)~\textbf{References}, an ordered array of reference objects, each grouping one source document with the constraints extracted from it (\autoref{fig:ref-schema}).
Constraint values are polymorphic: scalars, ranges $\{\min,\max\}$, algebraic expressions, or keyed look-up tables---enabling faithful encoding of diverse regulatory formats.
A single generated database can therefore aggregate references from international codes (e.g.\ IRC~2021~\cite{irc2021}), national space standards (e.g.\ NDSS~2015~\cite{ndss2015}), and domain-specific guidelines (e.g.\ NHS~HBN~03-01~\cite{nhs-hbn0301}) in one unified file.

\begin{figure}[t]
\centering
\begin{lstlisting}[basicstyle=\ttfamily\scriptsize,frame=single]
Reference
+-- ref_id        : unique slug
+-- building_type : residential | office | hospital
+-- jurisdiction  : geographic / regulatory context
+-- retrieved_at  : ISO-8601 date
+-- sources[]
|   +-- source_id, type, title, publisher, url
+-- constraints[]
    +-- constraint_id, name
    +-- constraint_type : minimum | maximum | range
        | ratio | recommendation | rule_of_thumb
        | table
    +-- value  : number | {min,max} | {expr} | table
    +-- unit   : m^2 | ft^2 | m | count | other
    +-- measurement_basis, applicability
    +-- confidence : high | medium | low
    +-- source_ids[] : provenance links
\end{lstlisting}
\caption{Simplified reference schema in the domain database.}
\label{fig:ref-schema}
\end{figure}

\subsubsection{Robot Infrastructure}\label{sec:db-robot}

Because \emph{robot placement, charging, circulation, and docking} is a first-class research topic (topic~h), every generated database contains robot-specific constraints alongside conventional architectural standards.
Typical automatically-retrieved constraints include:
charging-dock minimum clearance radius (e.g.\ ${\geq}\,1.2$\,m for 180\textdegree{} approach),
robot staging-zone area per unit (e.g.\ 2.5\,m$^2$),
corridor width minimums for bidirectional robot traffic (e.g.\ ${\geq}\,1.8$\,m),
and line-of-sight requirements between robot zones and staff stations for safety monitoring.
These constraints are injected into generation prompts alongside dimensional and safety rules, ensuring that produced floorplans are robot-ready without post-hoc modification.

\subsubsection{Constraint Retrieval}\label{sec:db-retrieval}

At floorplan generation time, the RAG retriever selects constraints from the generated database in two steps:
\begin{enumerate}
  \item \textbf{Building-type classification.}
    Room-type labels from the input scene graph are matched against keyword sets (e.g.\ \textit{patientroom}, \textit{operatingroom} $\to$ hospital; \textit{meetingroom}, \textit{conferenceroom} $\to$ office; default $\to$ residential).
  \item \textbf{Constraint filtering and formatting.}
    All references whose building type matches the classified type are selected.
    Their constraints are rendered as natural-language bullets and injected into the LLM prompt under a \emph{Reference Room Size Standards} heading.
\end{enumerate}
This two-stage retrieval ensures that the generative model receives only domain-relevant constraints, keeping the prompt concise while maximizing compliance with the standards discovered during database generation.

\subsection{Generation Pipeline}
Our generation pipeline is multi-staged, leveraging the domain database in combination with the asset database for RAG in order to generate floorplans and object placements. Before we explain the full pipeline, we first define the asset database:

\setcounter{subsubsection}{0}
\subsubsection{Asset Database}
The asset database is an automatically processed dataset of 3D models (multi-format, mainly USDZ) that is queryable via natural language.

The current database spans three domains with the following asset counts:
\begin{center}
\begin{tabular}{lcccc}
\toprule
\textbf{Domain} & \textbf{Office} & \textbf{Hospital} & \textbf{Residential} & \textbf{Total} \\
\midrule
\textbf{\# 3D Assets} & 139 & 71 & 49 & \textbf{259} \\
\bottomrule
\end{tabular}
\end{center}

For each asset, the database computes and stores the following annotations: 
\begin{itemize}
  \item \textbf{Bounding box}: axis-aligned 3D extents
        $[x_{\min}, x_{\max}] \times [y_{\min}, y_{\max}] \times [z_{\min}, z_{\max}]$ in meters;
  \item \textbf{Material list}: physical surface materials
        (e.g.\ leather, steel, wood);
  \item \textbf{Color palette}: dominant colors
        (e.g.\ white, silver);
  \item \textbf{Human--object interactions (HOI)}: affordance tags
        (e.g.\ lie, sit, operate);
  \item \textbf{Face direction}: canonical front-facing axis
        ($+x$, $-x$, $+y$, $-y$, $xy$) for consistent placement orientation;
  \item \textbf{Semantic tags}: domain and category labels
        (e.g.\ office, hospital, bedroom, furniture);
  \item \textbf{Free-text description}: a short natural-language
        description (e.g.\ \emph{Gurney}, \emph{Armchair}).
\end{itemize}

All annotations are concatenated into a single text embedding, enabling nearest-neighbor retrieval for arbitrary natural-language asset queries (e.g.\ querying \enquote{adjustable hospital bed} returns the closest matching assets).

\subsubsection{Text-to-Graph}
Given a natural-language prompt (e.g.\ \enquote{Design an office floorplan for 60 staff}), the \emph{Text-to-Graph} module generates a \emph{scene graph} $G = (V, E)$ where vertices $V$ are rooms and edges $E$ encode adjacency and door relationships.  The module uses a system prompt that specifies a strict JSON schema for nodes and edges, and provides few-shot in-context examples loaded from domain-specific supporting-example files. 

Each vertex carries:
\begin{itemize}
  \item A unique room identifier
  \item A semantic room-type label
        (e.g.\ \emph{Reception}, \emph{Patient Room})
  \item A preliminary list of 8--15 asset names per room,
        enriched by an \emph{asset enrichment} pass that fills gaps using room-type keyword hints (e.g.\ robot rooms receive charging docks, robot carts, maintenance benches)
  \item A natural-language room description
\end{itemize}

Beyond room labels, the scene graph also includes \emph{robot-operation semantic annotations} attached to nodes and edges.
Concretely, the parser creates typed regions such as \texttt{cleaning\_zone}, \texttt{service\_zone} and \texttt{charging\_buffer}, then links them to corridor/door edges that robots should traverse.
For example, cleaning robots in office domains are assigned high-coverage cleaning zones in open work areas and low-priority zones in meeting rooms, while service robots in hospital domains receive service zones around nursing stations, patient-room doors, and supply rooms.
These annotations are stored as role-tagged constraints and propagated to floorplan and placement prompts.

\subsubsection{Floorplan Generation}

The scene graph is passed to the floorplan generator, which implements a three-phase RAG pipeline: 
    \paragraph{Preprocess} The generator retrieves the building type and the RAG database from the Domain Database Pipeline. Based on the building type, the system retrieves relevant constraints from the knowledge base, which are formatted as natural language and injected into the LLM prompt.
    
    \paragraph{Inference} The structured prompt, with the RAG context injected, is submitted to the LLM, which returns a JSON object containing polygon coordinates for rooms and doors.

    \paragraph{Post-processing} The post-processing stage parses this response, constructs Shapely~\cite{shapely} geometries, and assembles the final \texttt{WorldDescription}. Wall segments are derived by computing boundary differences between room polygons and door openings. Each zone receives material assignments based on room type.

\begin{figure*}[b]
\centering
\includegraphics[width=0.99\linewidth]{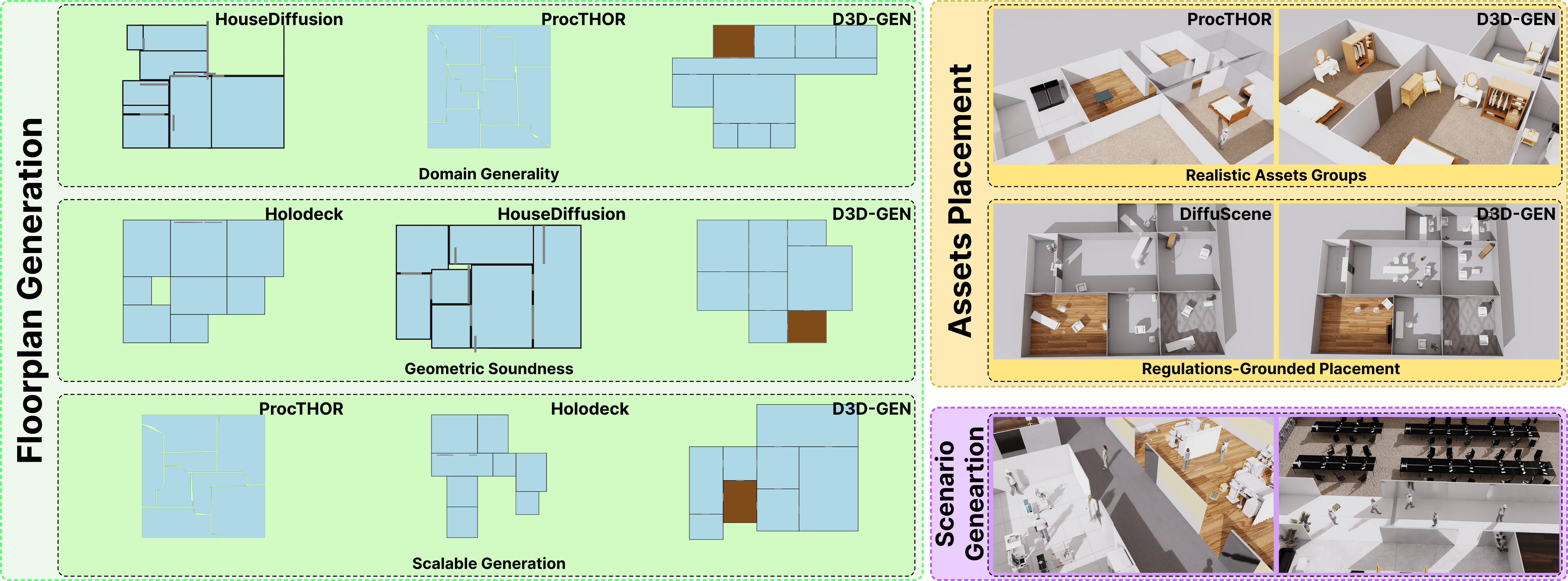}
\par\smallskip
\caption{%
    Qualitative comparisons. The same prompt
    is used for generating scene graphs, across three domains (Hospital, Residential, Office). 
    For the \emph{Floorplan Generation} task, the same scene graph is used for each baseline compared with \WORKNAME. \WORKNAME excels on failure modes of baselines (wrong or no positioning of doors, coordinates not according to actual rules, scalability with increasing number of rooms).
    The same output of the Floorplan Generation pipeline \texttt{WorldDescription} object is used for the \emph{Asset Placement} task. \WORKNAME has better object handling and grouping, as well as better knowledge of assets rotation with respect to walls and room boundaries. Robot charging rooms and clearance zones were designated only by \WORKNAME, marked in brown.
    Scenario Generation outputs Human-Robot interaction scenarios where human presence provides robots with more context in worlds.
}
\label{fig:baselines-comparison}
\end{figure*}

\subsubsection{Asset Population}

After the floorplan is generated, it is passed to a three-part \emph{population pipeline}. Given a room polygon, door geometry, and semantic room label from the floorplan stage, the module outputs a set of placed objects with collision-free 2D poses. 

    \paragraph{Asset retrieval} The pipeline obtains the building type from the floorplan generation pipeline. Based on the building type, the system maps the scene graph assets to real models via multi-stage matching. The room-type priors are added when requested assets are sparse. Candidate ranking combines lexical relevance and metadata compatibility by considering text similarity, room-type compatibility, and asset sizes.
    
    \paragraph{LLM-guided placement proposal} The placement model receives:
    \begin{itemize}
      \item room boundaries (corners and bounding box),
      \item door polygons and clearance instructions,
      \item ranked candidate assets with nominal dimensions,
      \item target assets count and room semantics.
    \end{itemize}
    Then it outputs a JSON list of proposed placements, which are validated by hard constraints (no overlapping, compliance with room boundaries, wall offsets). If the proposal fails, the system performs position search to place the assets, subject to all constraints.
    
    \paragraph{Output canonicalization} Accepted placements are exported as obstacle entities with unique object identifiers and domain-relative model paths. The populated room list is then serialized into the final world artifact \texttt{WorldDescription} aligning with the generated floorplan graph.

Finally, the RAG pipeline exports our world format as a \texttt{world.yaml} file. Each zone includes: (1)~geometry as corner positions, (2)~walls with material specs, (3)~doors with width/pose, and (4)~semantic name (e.g., \texttt{patient\_room\_01}), and (5) populated assets (e.g., \texttt{Hospital/Patient\_Bed\_A}). This ensures ROIs/POIs in behavior specifications are grounded in geometrically valid, standards-compliant regions.

\section{Evaluations}

We evaluate \WORKNAME on its ability to (1) reproduce plausible results in common domains, and (2) extend to less-common transfer domains. To this end, we chose \emph{indoor residential} \textbf{(Residential)} because of its ubiquity as the common domain, \emph{commercial office space and hospital space} \textbf{(Office, Hospital)} as transfer domains.

We generated 450 worlds in total: 150 residential, 150 office, and 150 hospital (50 samples each for requested room counts 5, 8, and 10). Quantitative metrics are parsed from \texttt{world.yaml}, and room polygon areas are computed with the Shoelace formula.

\begin{table}[!htp]
\centering
\caption{Domain-agent execution profile. \enquote{MTok} denotes million tokens (input + output) across all LLM calls.}
\label{tab:agent-exec}
\scriptsize
\setlength{\tabcolsep}{2.5pt}
\renewcommand{\arraystretch}{0.95}
\begin{tabularx}{\columnwidth}{X*{8}{c}}
\toprule
\textbf{Domain} & \textbf{In} & \textbf{Out} & \textbf{Total} & \textbf{Infer} & \textbf{Research} & \textbf{Struct.} & \textbf{Synth.} & \textbf{Total (s)} \\
 & \textbf{(MTok)} & \textbf{(MTok)} & \textbf{(MTok)} & \textbf{(s)} & \textbf{(s)} & \textbf{(s)} & \textbf{(s)} &  \\
\midrule
Residential & 0.42 & 0.18 & 0.60 & 2.1 & 148.5 & 31.4 & 1.8 & 183.8 \\
Office & 0.51 & 0.22 & 0.73 & 2.3 & 162.7 & 38.2 & 2.1 & 205.3 \\
Hospital & 0.56 & 0.24 & 0.80 & 2.2 & 174.3 & 42.6 & 2.3 & 221.4 \\
\bottomrule
\end{tabularx}
\end{table}

\begin{table}[!htp]
\centering
\caption{Generated database quality and size metrics.}
\label{tab:agent-db}
\scriptsize
\setlength{\tabcolsep}{2.5pt}
\renewcommand{\arraystretch}{0.94}
\begin{tabularx}{\columnwidth}{Xccc}
\toprule
\textbf{Metric} & \textbf{Residential} & \textbf{Office} & \textbf{Hospital} \\
\midrule
Raw notes (chars) & 86,400 & 104,200 & 80,788 \\
Chunk count & 8 & 9 & 7 \\
Corpus documents (dedup) & 24 & 29 & 41 \\
Final references & 24 & 29 & 41 \\
Total constraints & 138 & 172 & 160 \\
Avg.\ constraints / reference & 5.8 & 5.9 & 3.9 \\
Unique room types covered & 12 & 16 & 82 \\
Constraint-type families & 4 & 5 & 4 \\
Jurisdiction coverage (count) & 3 & 4 & 9 \\
Confidence split (H/M/L) & 42/48/10\% & 38/51/11\% & 90/9/1\% \\
DB file size (KB) & 86 & 112 & 116 \\
\bottomrule
\end{tabularx}
\end{table}

\begin{table*}[t]
\centering
\caption{Multi-method comparison on generated outputs. \textbf{FP} is floorplan generation-only methods, \textbf{P} is population-only methods, \textbf{FP+P} combines both. Quantitative metrics are extracted from automated world parsing; perceptual metrics are Gemini-based averages across domains. For Layout-FID, lower is better; for Layout, Visual, and VQA Acc., higher is better.}
\label{tab:multi-method-results}
\scriptsize
\setlength{\tabcolsep}{3pt}
\begin{tabular}{lccccccccccc}
\toprule
Method & $n$ & \multicolumn{6}{c}{Quantitative Metrics} & \multicolumn{4}{c}{Perceptual Metrics} \\
\cmidrule(lr){3-8}\cmidrule(lr){9-12}
 &  & Objects & Obj./Room & Doors & Area (m$^2$) & Obj. Spacing (m) & Layout-FID & Layout & Visual & VQA Acc. & Scene Rating (P) \\
\midrule
HouseDiffusion (FP) & 18 & - & - & 8.31 & 40.5 & - & 52933.82 & 4.48 & 4.60 & 0.313 & - \\
Holodeck (FP) & 18 & - & - & 7.00 & 154.4 & - & 12812.76 & \textbf{8.46} & 7.17 & 0.353 & - \\
ProcTHOR (FP+P) & 18 & 3.73 & 0.52 & 8.11 & 938.7 & 6.41 & 85517073.63 & 5.50 & 5.78 & 0.333 & 8.09 \\
DiffuScene (P) & 18 & 46.60 & 6.33 & - & - & 8.58 & \textbf{2531.05} & 7.78 & 7.22 & 0.389 & 8.33 \\
D3D-GEN (FP+P) \textbf{(Ours)} & 18 & \textbf{54.56} & \textbf{7.06} & 7.67 & 244.3 & \textbf{9.19} & 5377.19 & 8.33 & \textbf{7.61} & \textbf{0.400} & \textbf{8.63} \\
\bottomrule
\end{tabular}
\end{table*}

\subsection{Quantitative Metrics}
\autoref{tab:eval-config-summary} compresses all per-instance quantitative data into 9 configuration rows (domain $\times$ requested room count).

\autoref{tab:multi-method-results} benchmarks \WORKNAME against four baselines spanning floorplan-only (\textbf{FP}: HouseDiffusion, Holodeck), population-only (\textbf{P}: DiffuScene), and combined (\textbf{FP+P}: ProcTHOR) generation paradigms. Each method produced $n{=}18$ worlds from the same residential prompts. We report six quantitative metrics extracted automatically from the generated \texttt{world.yaml} files and four perceptual metrics obtained via Gemini-based evaluation. \WORKNAME achieves the highest object count, object density, object spacing, visual quality, VQA accuracy, and Scene rating (derived from~\cite{tam2025sceneeval}), while remaining competitive on Layout-FID. This demonstrates that domain-grounded RAG generation produces worlds that are both structurally richer and perceptually more realistic than existing approaches.

\begin{table}[t]
\centering
\caption{Configuration-level quantitative summary from generated worlds (n=50). Values are means per configuration.}
\label{tab:eval-config-summary}
\scriptsize
\setlength{\tabcolsep}{4pt}
\renewcommand{\arraystretch}{0.95}
\begin{tabularx}{\linewidth}{>{\hsize=1.8\hsize\raggedright\arraybackslash}X*{3}{c}*{2}{>{\centering\arraybackslash\hsize=0.6\hsize}X}}
\toprule
\textbf{Domain} & \textbf{Objs} & \textbf{Doors} & \textbf{Walls} & \textbf{Area~(m$^2$)} & \textbf{Objs./Room} \\
\midrule
Residential (5-room)  & 30 & 5  & 30 & 95  & 6.1 \\
Residential (8-room)  & 48 & 9  & 50 & 158 & 6.0 \\
Residential (10-room) & 60 & 11 & 62 & 222 & 6.0 \\
\addlinespace
Office (5-room)       & 45 & 6  & 32 & 287 & 8.9 \\
Office (8-room)       & 64 & 9  & 49 & 347 & 7.9 \\
Office (10-room)      & 72 & 10 & 60 & 365 & 7.2 \\
\addlinespace
Hospital (5-room)     & 35 & 5  & 30 & 114 & 7.0 \\
Hospital (8-room)     & 56 & 8  & 48 & 231 & 7.0 \\
Hospital (10-room)    & 69 & 10 & 61 & 246 & 6.9 \\
\bottomrule
\end{tabularx}
\end{table}

\subsection{Qualitative Comparison with Baselines}

Our qualitative analysis focuses on commonly observed failure modes and how our system successfully avoids them, yielding more consistent, higher-quality results.
\autoref{fig:baselines-comparison} compares \WORKNAME side-by-side with HouseDiffusion~\cite{housediffusion}, ProcTHOR~\cite{procthor}, Holodeck~\cite{holodeck}, and DiffuScene~\cite{diffuscene} on two main tasks: Floorplan Generation and Asset Placement.
\subsubsection{Floorplan}
We observe three consistent floorplan-level patterns across configurations.
First, room-count fidelity is perfect: in every case, the realized number of rooms exactly matches the requested count.
Second, total area scales differently by domain: office and hospital worlds expand more substantially than residential layouts, reflecting larger functional rooms and greater circulation requirements.
Third, topological complexity is highest in hospitals, which exhibit the largest numbers of doors and wall segments, consistent with corridor-loop structures and care-workflow constraints.
Beyond these aggregate trends, qualitative inspection also shows that layouts generated by \WORKNAME maintain clearer adjacency logic and more coherent transitions between private and shared spaces, which is particularly important for social navigation tasks.
We also observe that robot-path semantics remain consistent with domain intent: office scenes tend to produce broad cleaning loops through open workspaces and circulation spines, while hospital scenes prioritize service routes that connect nursing stations, patient-room thresholds, and utility rooms with fewer unnecessary detours.

\subsubsection{Asset Population}
At the population level, two additional patterns are consistent.
Object density varies strongly by domain: office environments are the most cluttered (up to $8.9$ objects per room in our configuration summary), whereas residential scenes remain comparatively sparse.
Material usage also differs by domain: hospital scenes show the least material diversity, while residential and office environments use broader material palettes.
In addition, object groupings in \WORKNAME are typically more functionally consistent with room semantics (e.g., workstation clusters in offices and treatment-support groupings in hospitals), while preserving usable free space for movement.
This carries over to robot-operational semantics: cleaning robots are assigned larger cleanable areas (open-plan office regions, lobby loops), whereas service robots are assigned targeted service areas near delivery-relevant assets (beds, nurse counters, supply shelves), reducing cross-traffic with human circulation.
This balance between semantic richness and clearance helps maintain both visual plausibility and simulation readiness.


\section{Conclusion}
In this paper, we introduced \WORKNAME as a novel approach to simulable 3D World Generation.
Unlike prior work, we take a fully user-defined domain approach that extracts a domain database from publicly accessible data. We actively include robot-semantic annotations to support human-robot interaction across contextually relevant use cases.

We have generated and evaluated 450 different worlds to show the scalability of \WORKNAME.
Additionally, we qualitatively compared our results to established baselines.
These results confirm \WORKNAME's applicability across both common and less-explored domains.

The main limitation is the remaining reliance on an existing asset database, which may not be readily available for every domain.
Infinigen-Articulated~\cite{infinigen-articulated} procedurally generates articulated assets for the \emph{indoor residential} domain.
In our future work, we will integrate a similar approach and extend it to arbitrary domains by analyzing asset data with our domain agent.
We already have the infrastructure to export just-in-time generated assets with the final world output.


Currently, the only direct interaction with the domain database is its initial generation from the prompt. The user might not be satisfied with the captured intent and the resulting database.
Another future extension is the modification of existing databases by prompting the domain agent.
This would allow the user to interactively generalize or specialize the knowledge graph until the domain matches their intent.

\section{Acknowledgement}
This paper is supported in part by the Ministry of Education (MOE) Academic Research Fund (AcRF) Tier 2 grant (Grant ID: T2EP20124-0055)

\bibliography{IEEEabrv,references.bib}

\end{document}